\documentclass{article}

\usepackage[preprint]{neurips_2026}

\usepackage[utf8]{inputenc} 
\usepackage[T1]{fontenc}    
\usepackage{hyperref}       
\usepackage{url}            
\usepackage{amsfonts}       
\usepackage{nicefrac}       
\usepackage{microtype}      
\usepackage{xcolor}         
\usepackage{amsmath} 
\usepackage{graphicx}
\usepackage{mathtools}
\usepackage{multirow} 
\newcommand{\myparagraph}[1]{%
  \par\vspace{-0.2em}%
  \noindent\textbf{#1}\hspace{0.3em}%
}
\DeclareMathOperator*{\argmax}{argmax}
\newcommand{\ie}{i.e.,\ }

\title{Tracing Generated Samples to Training-Data Clusters in Flow-Matching Models}
\title{Tracing Generated Samples to Training-Data Clusters in Flow-Matching Models}

\author{%
  Rania Briq\\
  Forschungszentrum J\"ulich\\
  Technical University Dortmund\\
  \texttt{r.briq@fz-juelich.de}
  \And
  Ohad Fried\\
  Reichman University
  \And
  Michael Kamp\\
  Technical University Dortmund\\
  Lamarr Institute\\
  Institute for AI in Medicine, University Hospital Essen
  \And
  Stefan Kesselheim\\
  Forschungszentrum J\"ulich\\
  Helmholtz AI\\
  University of Cologne
}

\begin{document}

\maketitle

\begin{abstract}
Understanding which training samples influence a generated image is an important problem in generative modeling. In flow matching, training samples influence the generated image through the velocity field along the generation trajectory. Removing samples to examine their counterfactual influence changes the velocity field, and the resulting effect on the final image depends on how the change propagates through the trajectory. Consequently, local changes in the velocity field do not necessarily predict the final counterfactual effect.

This work investigates attribution in flow-matching models through a hybrid analytical--learned approach, and uses it to derive trajectory-based attribution scores at the cluster level. 
We evaluate these attribution scores using independently retrained leave-one-cluster-out (LOCO) models, and compare with several attribution baselines using two different flow-matching latent spaces. Our experiments show that semantic similarity constitutes a strong baseline, while the closed-form trajectory-based attribution is competitive in some metrics without requiring counterfactual retraining or model gradients.

Our results show that attribution in flow matching depends not only on semantic similarity to training samples, but also on the latent representation, trajectory dynamics, and how influence is propagated to the final output. 
\end{abstract}
\section{Introduction}

Attribution is a central problem in generative modeling, that helps explain generated content and interpret model behavior. It also has societal implications since it can address copyright problems and help mitigate unwanted behavior. The question of attribution can be investigated in multiple ways, for example, determining which training samples resemble a generated output, or determining the counterfactual effect, \ie answering how removing a set of samples would change the generated image. Here, we study training-data attribution in flow matching (FM) at the level of training-data clusters, by determining the counterfactual effect of removing clusters, and comparing with various baselines that are also based on semantic similarity. We evaluate the various approaches based on how well they predict the magnitude of the counterfactual change. 

Flow matching generates samples by integrating a learned velocity field along a trajectory. Data removal affects not only the endpoint, but also the entire flow trajectory and its dynamics. For instance, a large local velocity perturbation does not necessarily imply a large change in the final output, because the change also depends on how the perturbation is propagated to the endpoint, where changes can cancel out and it becomes computationally expensive to track them. The flow matching formulation has an analytical closed-form (CF) solution of the optimal velocity field~\citep{gao2024flow,bertrand2025closed}, which takes the form of a linear combination of contributions from each training sample where a weight factor describes the magnitude of each sample's contribution, therefore its structure provides an ideal way to study attribution and predict the counterfactual effect. The approach evaluates the contribution of clusters to the velocity field along the trained model's trajectory, making it an analytical and model-based hybrid approach. It can dynamically adapt to new clusters, while not requiring any retraining or finetuning. 

Our study is partly motivated by the observation that latent flow-matching models are very stable under training-sample removal: only when a semantic group loses its support by dropping its corresponding cluster completely does the generated sample's identity change \citep{briq2026exploring}. This indicates that individual samples may have weak influence in isolation, while a cluster has a more distinguishable effect. We accordingly partition the training data into clusters, and for each cluster train a flow-matching model while leaving that cluster out. These leave-one-cluster-out (LOCO) models are then used as counterfactual oracles for evaluating attribution scores. 

Our experiments are conducted using two different flow-matching latent representations on the CelebA-HQ dataset~\citep{karras2017progressive}. The closed-form trajectory-based attribution reflects the whole trajectory and is FM-interpretable. Our observations reveal that attribution in flow matching is not explained by endpoint similarity or local velocity-field perturbations alone. Rather, attribution depends on whole-trajectory dynamics, the counterfactual metric involved (e.g. endpoint-latent deviation or velocity-field perturbation), and the flow-matching latent representation.

Existing methods for attribution include influence functions~\citep{koh2017understanding}, which approximate the effect of removing certain training samples on the parameters. Other methods include gradient-based methods such as TracIn~\citep{pruthi2020estimating,park2023trak}. These methods have also been adapted to diffusion models, including timestep-aware gradient attribution~\citep{georgiev2023journey,xie2024data}, influence-function formulations~\citep{zheng2024intriguing,mlodozeniec2025influence}. Related to our cluster-level setting, GUDA~\citep{murata2026guda} studies group-level attribution through approximate counterfactual unlearning of the likelihood. Unlike FM-CF, adapting to new clusters requires retraining. Analogously to FM-CF, NDA~\citep{zhao2025nonparametric} derives nonparametric patch-level attribution using the analytical optimum of the diffusion objective. In contrast to FM-CF, it is completely model-free and therefore cannot distinguish between the various trained models and their dynamics or training stochasticity.

\begin{figure*}[t]
    \centering
    \includegraphics[width=\textwidth]{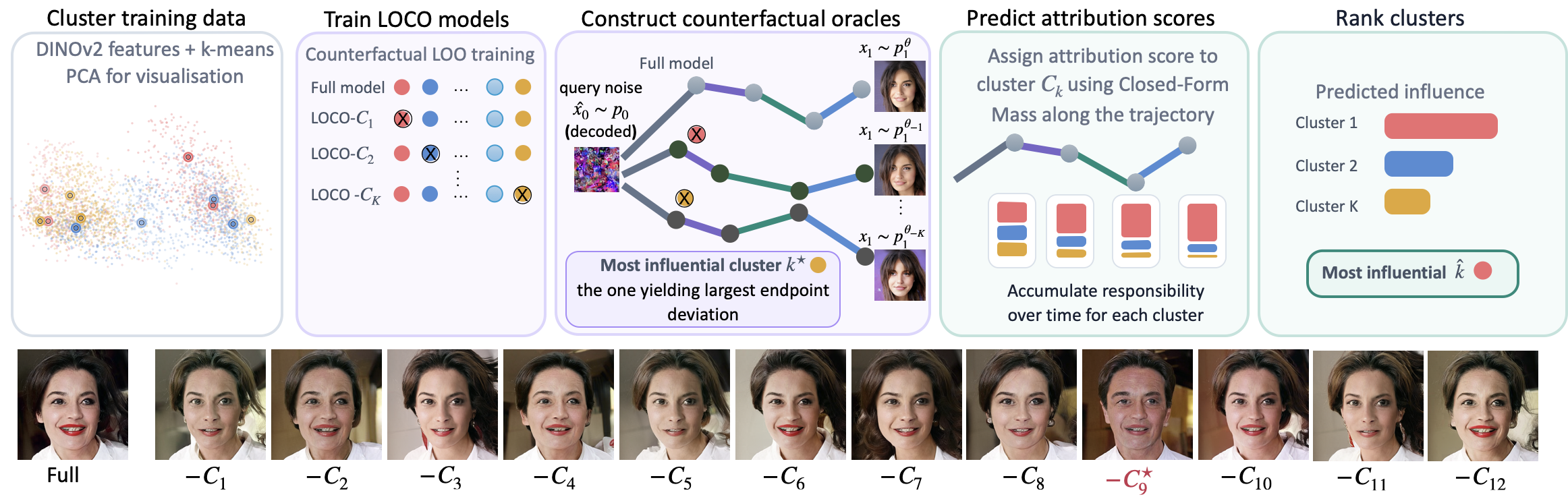}
    \caption{\textbf{Top:} Overview of the proposed attribution approach based on rectified-flow closed-form solution (CF). We train counterfactual leave-one-cluster-out (LOCO) models for evaluation. CF-based attribution assigns an attribution score to each cluster by evaluating its contribution along the trajectory. \textbf{Bottom:} Full-model output and $K=12$ counterfactual images generated by the LOCO models. $-C_k$ denotes the model retrained without cluster $C_k$. $C_9$ yields the highest RMS, making it the winning (most influential) cluster.}
    \label{fig:approach}
\end{figure*}

To summarize, our contributions are as follows. (i) We introduce cluster-level training-data attribution in flow-matching models and propose a cluster-level attribution setting defined by leave-one-cluster-out (LOCO) models. (ii) We derive an attribution approach that combines the analytical FM closed-form solution with the trained model trajectory. (iii) We evaluate against existing attribution baselines such as gradient-based approaches and semantic similarity. An overview of the approach is illustrated in Fig.~\ref{fig:approach}.

\section{Methodology}
\myparagraph{Flow Matching preliminaries.} We consider the rectified flow formulation which uses linear interpolants and samples $x_0\sim p_0=\mathcal N(0,I)$ and $x_1\sim p_1$ independently from the source and target distributions $p_0$ and $p_1$. The conditional probability path and its corresponding path velocity are
\begin{equation}
    x_t=(1-t)x_0+t x_1,
    \qquad
    u_t(x_0,x_1)=x_1-x_0,
    \qquad t\in[0,1].
    \label{eq:rectified_path}
\end{equation}
A velocity field $v_\theta(x,t)$ is learned by minimizing the objective
\begin{equation}
\mathcal L(\theta)
=
\mathbb E_{\substack{
x_0\sim p_0,\;x_1\sim p_1,\\
t\sim\mathcal U[0,1]
}}
\left[
\left\|
v_\theta(x_t,t)-(x_1-x_0)
\right\|_2^2
\right].
\label{eq:fm_objective}
\end{equation}
The optimal regression function is $v^*(x,t)=\mathbb E[u_t\mid x_t=x]$. At inference, samples are generated by integrating $\dot{x}_t=v_\theta(x_t,t)$ from $x_0\sim p_0$ to obtain $x_1 \sim p_1$. Following common FM models, our flow-matching models operate in the latent space of an autoencoder~\citep{ma2024sit,zheng2026diffusion}.

\myparagraph{Closed-form optimal velocity for Rectified Flow.} The minimizer of Eq.~\ref{eq:fm_objective} is the conditional expectation of the target velocity~\citep{gao2024flow,bertrand2025closed} that yields for a finite dataset
\begin{equation}
		\begin{aligned}
			v_{\mathrm{CF}}(x,t)\coloneqq\hat{u}^*(x,t) &= \sum_{i=1}^{n} \lambda_i(x,t)\,\frac{x^i-x}{1-t},\quad
			\lambda_i(x,t) &= \operatorname{softmax}_i\!\left(-\frac{\lVert x-tx^{i}\rVert^2}{2(1-t)^2}\right),
		\end{aligned}
		\label{eq:vf_closedform}
	\end{equation}
where $n$ denotes the dataset size. We can interpret the strictly positive coefficient $\lambda_i(x,t)$ as the influence of sample $i$ at point $x$ for a given timestep $t$. 
This analytical solution gives rise naturally to an attribution method along the model's learned trajectory $x_t^{\theta}$. For each timestep $t$, we evaluate the closed-form velocity by substituting $x=x_t^{\theta}$.
It is noteworthy that the network does not reproduce the exact optimal velocity field, otherwise it would just reproduce the training samples rather than generate novel samples~\citep{gao2024flow,bertrand2025closed}.

\myparagraph{Trajectory-based cluster-level attribution}
For our cluster-based attribution scheme, we assume the data is partitioned into clusters, and the task is to decide which cluster has the strongest influence on a given generated sample. For our experiments, given a training dataset $\mathcal{D}$, we cluster its samples into $K$ clusters using k-means based on the cosine distance in DINOv2 embedding space. 
For an intermediate state $x_t^\theta$ generated by the model, we define the posterior mass assigned to cluster $C_k$ as 
\begin{equation}
\Lambda_k(x_t^\theta,t)
=
\sum_{i\in C_k}
\lambda_i(x_t^\theta,t).
\label{eq:cluster_posterior_mass}
\end{equation}
Since we are interested in a cluster's influence along the whole trajectory, we define the trajectory-based attribution score assigned to a cluster as the integral:
\begin{equation}
A_k^{\mathrm{mass}}
=
\int_0^1
\Lambda_k(x_t^\theta,t)dt,
\label{eq:integrated_cluster_mass}
\end{equation}
\ie the cumulative softmax weights each cluster contributes to the velocity field along the trajectory. The most influential cluster is predicted to be the maximum-score cluster: $\hat{k}= \argmax_{k}A_k^{\mathrm{mass}}$. We refer to this method as \emph{Closed-form cluster mass}.

The posterior-mass quantity in Eq.~\ref{eq:cluster_posterior_mass} quantifies how much a given sample cluster contributes to a velocity field at a specific point, but it does not tell us how much this cluster's removal perturbs the velocity field. We therefore introduce another CF-based scoring variant that selects the cluster $k$ whose deletion disturbs the velocity field the most. Concretely, we measure how much removing cluster $C_k$ perturbs the local velocity field using
\begin{equation}
v_{\mathrm{CF}}^{-k}(x,t)
=
\frac{
\displaystyle
\sum_{i\notin C_k}
\lambda_i(x,t)
\frac{x^i-x}{1-t}
}{
1-\Lambda_k(x,t)
},
\label{eq:deleted_closed_form_velocity}
\end{equation}
where the denominator $1-\Lambda_k(x,t)$ is added to renormalize the softmax weights after removing $C_k$.
We aggregate these local perturbations along the whole trajectory by integrating 
\begin{equation}
A_k^{\mathrm{del}}
=
\int_0^1
\frac{
\left\|
\delta v_{\mathrm{CF}}^k (x_t^\theta,t)
\right\|_2
}{
\sqrt{d}
}
\,dt,
\qquad
\delta v_{\mathrm{CF}}^k (x_t^\theta,t)
=
v_{\mathrm{CF}}(x_t^\theta,t)
-
v_{\mathrm{CF}}^{-k}(x_t^\theta,t).
\label{eq:integrated_deletion_attribution}
\end{equation}

where $d$ is the latent dimensionality, and $v_{\mathrm{CF}}^{-k}$ is the velocity field using all clusters except $C_k$. The integral is approximated using the trapezoidal rule. For attribution, we select the cluster that yields the largest perturbation. In practice, we find that it performs nearly equally to the posterior mass while requiring a more expensive computation.

\myparagraph{The oracle counterfactuals.}
To evaluate whether the attribution scores actually reflect a retrained model's behavior, we create counterfactual models by performing leave-one-cluster-out (LOCO) training. Specifically, for each cluster $C_k$, we train a counterfactual model on $\mathcal{D}^{-k}=\mathcal{D}\setminus C_k$.

For a given query $z$, let $x_{1,z}^{\theta}$ denote the endpoint generated by the full model and $x_{1,z}^{\theta_{-k}}$ the endpoint generated by the model trained without cluster $C
_k$. For each LOCO model, we evaluate the distance between its output and that of the model trained on the full dataset. This root mean square distance (RMS) is given by 
\begin{equation}
E_k
=
\frac{
\left\|
x_{1,z}^{\theta_{-k}}
-
x_{1,z}^\theta
\right\|_2
}{
\sqrt{d}}.
\label{eq:counterfactual_rms_gt}
\end{equation}
We treat the cluster whose LOCO produces the largest endpoint deviation as the most influential cluster: $k^{\star}=\arg\max_k E_k$, which constitutes our pseudo-ground-truth. Visually, this is shown in Fig~.\ref{fig:approach} (bottom), where the LOCO model of $C_9$ creates the largest endpoint deviation, consistent with the observation that the generated image is the most dissimilar. 

Finally, to reduce the variance stemming from training stochasticity, we train $S=5$ counterfactual models per deleted cluster and average $E_k$ over these seeds.

\section{Experiments}
\myparagraph{Latent representations.} 
We evaluate attribution using two FM variants, each trained and operating in a different latent representation space. Specifically, DINOv2-based Representation Autoencoder (RAE)~\citep{oquab2023dinov2}, and a variational autoencoder (VAE) using SiT framework~\citep{ma2024sit}. The motivation is to test to what extent attribution depends on the geometry of the latent space, since attribution scores based on the closed-form solution include computing distances between latent representations. In particular, we examine whether attribution based on semantic similarity might favor the more semantically structured DINOv2 space over the VAE's space.

\myparagraph{Dataset and training.} We perform our experiments on the CelebA-HQ dataset, which contains high-resolution celebrity faces~\citep{karras2017progressive}, and use its 28k training split for training. Both the RAE and SiT models are trained for $700$ epochs using a batch size of 128 and the default training configurations of the respective frameworks.

\myparagraph{Clustering.} Since the RAE is DINOv2-based, we extract DINOv2 features of the training samples, apply mean-pooling spatially and cluster them into $K=12$ groups. We use these clusters to train the LOCO counterfactual models and to evaluate each cluster's contribution to a given query using the closed-form formula. In an ablative experiment, we also cluster the samples in CLIP embedding space~\citep{radford2021learning} and train the corresponding LOCO counterfactual models for further evaluation.

\myparagraph{Evaluation metrics}
For each query $z$, all counterfactual models are evaluated using the same initial noise as the full model. As defined above, let $\bar E_k(z)$ denote the seed-averaged counterfactual effect of deleting
cluster $C_k$, and let
\begin{equation}
 k^\star(z)=\arg\max_k \bar E_k(z)
 \label{eq:k_star}
 \end{equation}
denote the oracle cluster, which we treat as the pseudo-ground-truth. An attribution method, such as the closed-form method, assigns scores $A_k(z)$ to each of the $K$ clusters and predicts .
\begin{equation}
\hat{k}(z)=\arg\max_k A_k(z)
 \label{eq:k_hat}
 \end{equation}

Given $N$ queries, we evaluate attribution using three metrics. The first, \emph{Counterfactual effect}, measures the RMS between the full model's endpoint latent and that of LOCO-$C_{\hat{k}}$:
\begin{equation}
\frac{1}{N}\sum_{j=1}^N RMS(x_{1,z_j}^\theta-x_{1,z_j}^{\theta_{-\hat k}})
 \label{eq:RMS_counterfactual}
 \end{equation}

The second metric, \emph{Top-1 agreement}, measures how often a method's prediction of $\hat{k}(z)$ matches the counterfactual oracle's, \ie

\begin{equation}
\operatorname{Top1}=
\frac{1}{N}
\sum_{j=1}^{N}
\mathbf{1}\left\{
\hat{k}(z_j)=k^{\star}(z_j)
\right\}
\end{equation}

The third metric, \emph{Spearman rank correlation}, measures the correlation between the complete predicted ranking and oracle ranking
\begin{equation}
\rho_{\mathrm{S}}=
\frac{1}{N}
\sum_{j=1}^{N}
\operatorname{Spearman}\left(
\left(A_k(z_j)\right)_{k=1}^{K},
\left(\bar{E}_k(z_j)\right)_{k=1}^{K}
\right)
\label{eq:spearman_correlation}
\end{equation}

\myparagraph{Attribution methods and baselines.} We evaluate the CF-based approach against endpoint-based semantic similarity, which includes finding the nearest cluster centroid and nearest member in DINOv2 and CLIP embedding spaces using cosine similarity, and a gradient-similarity baseline. We additionally evaluate a random-cluster baseline, which we expect to have $1/K$ Top-1 agreement, to verify that the attribution scores are meaningful. 

\subsection{Results and Analysis}

\myparagraph{RAE-model evaluation.} 
We evaluate over $N=512$ generated query images. Table~\ref{tab:rae_main_single_seed} (left) compares attribution methods using the DINOv2-based RAE models while evaluating a single seed. It can be observed that attribution based on the endpoint semantic similarity is a strong baseline, such that DINOv2-centroid similarity achieves the highest counterfactual RMS and Top-1 agreement. Closed-form group mass achieves slightly worse attribution results and slightly better Spearman correlation, indicating that while it might not predict the most influential cluster accurately, it is better at predicting the overall ordering of influential clusters. Even though the nearest-member similarity iterates over all samples, it performs worse than nearest-centroid. Closed-form deletion and mass nearly perform equally. CLIP-based similarity inside DINOv2 clusters also achieves competitive performance, while slightly underperforming both closed-form and DINO. The strong performance of semantic-based similarity indicates it accounts for a big part of the attribution. Interestingly, none of the methods recover the oracle's counterfactual better than about 50\%. Hence, all these methods are missing additional factors.

\myparagraph{Training-seed sensitivity.} When we measured Top-1 pairwise training-seed variation across the oracle models, we observed that they agree only $25\%$ of the time, compared to $8.33\%$ random selection. Spearman correlations yield $\approx0.15$. This result indicates that counterfactual attribution is substantially affected by training stochasticity in addition to cluster removals, and motivated using the averaged counterfactual effect $\bar E_k$ over seeds as pseudo-ground-truth. We report the results in Table~\ref{tab:multi_seed_main_rae} (right). This result considerably reduces the gap between the oracle's counterfactual and DINOv2 centroid and closed-form as the oracle's RMS is considerably reduced from 0.79 to 0.71.

\begin{table}[t]
\centering
\small
\setlength{\tabcolsep}{4pt}
\begin{tabular}{lccc@{\hspace{10pt}}ccc}
\hline
&
\multicolumn{3}{c}{Single-seed oracle}
&
\multicolumn{3}{c}{Multi-seed oracle}
\\
\cline{2-4}
\cline{5-7}
Method
&
RMS $\uparrow$
&
Top-1\% $\uparrow$
&
Spearman $\uparrow$
&
RMS $\uparrow$
&
Top-1 \%$\uparrow$
&
Spearman $\uparrow$
\\
\hline
Random cluster
& 0.5926 & 8.33 & $\approx 0$
& 0.5833 & 8.33 & $\approx 0$
\\
Gradient similarity
& 0.6575 & 25.0 & 0.132
& 0.6507 & 33.3 & 0.286
\\
CLIP centroid
& 0.6725 & 27.9 & 0.183
& 0.6613 & 42.9 & 0.321
\\
CLIP nearest member
& 0.6789 & 27.9 & 0.208
& 0.6569 & 36.5 & 0.343
\\
Closed-form deletion
& 0.6811 & 29.7 & 0.223
& 0.6721 & 48.4 & 0.373
\\
Closed-form mass
& 0.6802 & 29.5 & \textbf{0.232}
& 0.6721 & 48.4 & \textbf{0.373}
\\
DINOv2 nearest member
& 0.6832 & 30.9 & 0.218
& --- & --- & ---
\\
DINOv2 centroid
& \textbf{0.6895} & \textbf{32.2} & 0.220
& \textbf{0.6816} & \textbf{51.6} & 0.369
\\
Oracle
& 0.7994 & 100.0 & 1.000
& 0.7179 & 100.0 & 1.000
\\
\hline
\end{tabular}
\caption{Attribution performance for DINOv2-based RAE models evaluated for  $N=512$ generated samples against single-seed and averaged multi-seed oracles. We omit nearest-member in multi-seed due to lack of clear advantage and the associated computational cost.}
\label{tab:rae_main_single_seed}
\label{tab:multi_seed_main_rae}
\vspace{-0.6em}
\end{table}
\myparagraph{SiT evaluation using the VAE latent space.} 
Similarly, to investigate how much attribution is dependent on the latent space, we perform the same evaluation using an SiT flow-matching model, which is based on a VAE latent space. Table~\ref{tab:vae_comparison} shows that DINOv2-based similarity is still competitive even though it is not the native representation of the model's latent space. This shows that DINOv2 provides a strong structured space, optimized to arrange images that share semantic or geometric similarities nearby. VAE centroid, which computes similarity to DINOv2-based cluster centroids in the latent space, underperforms all other baselines. Furthermore, in the VAE latent space, the overall attribution performance drops, indicating distances computed between semantically structured latents such as DINOv2 are more meaningful, whereas the VAE latent space is optimized for reconstruction and is less suitable for measuring similarity.

\begin{table}[t]
\centering
\begin{tabular}{lcc}
\hline
Method & Top-1 \%$\uparrow$ & Spearman $\uparrow$ \\
\hline
VAE centroid              & 18.8           & 0.118          \\
CLIP-sim to DINOv2 centroids & 28.1        & 0.175          \\
Closed-form mass          & 28.1          & 0.166          \\
DINOv2 centroid             & \textbf{34.4} & \textbf{0.223} \\
\hline
\end{tabular}
\caption{Attribution performance for the VAE-based flow-matching model. $N=128$.}
\label{tab:vae_comparison}
\vspace{-0.6em}
\end{table}

\section{CLIP-defined clusters}
To test whether DINOv2's strong performance stems from alignment between the latent representation of the FM model and DINOv2-based clusters being in the same space, we cluster the dataset in CLIP embedding space using $K=12$, and retrain the LOCO counterfactual models for each cluster. We then repeat the single-seed evaluation in Table~\ref{tab:rae_main_single_seed} with respect to these clusters, and report the results in Table~\ref{tab:rae_clip_defined_clusters}. It can be observed that the same pattern holds in terms of how well each method performs, with DINOv2-based similarity with respect to CLIP clusters still outperforming CLIP-based similarity. 

This result, taken together with the VAE space experiment, suggests that DINOv2's strong performance is not explained solely by the alignment between the representation that defines the attribution and that defines the clustering space or model's latent space. All methods achieve lower Top-1 and Spearman correlation compared to DINOv2-defined clusters, which could point that DINOv2's structured space organizes the training samples in a way that helps predict the counterfactual effect. For example, unlike CLIP, which focuses more on capturing semantic features by alignment with text features during its training, DINOv2 is trained to capture more geometric and lower-level features, which seems important for image similarity.
\begin{table}[ht]
\centering
\begin{tabular}{lccc}
\hline
Method & Selected RMS $\uparrow$ & Top-1 $\uparrow$ & Spearman $\uparrow$ \\
\hline
Random cluster        & 0.5865          & 8.3\%           & $\approx 0$    \\
Gradient similarity              & 0.6258          & 13.3\%          & 0.119          \\
Closed-form mass      & 0.6339          & 17.4\%          & 0.151          \\
Closed-form deletion  & 0.6351          & 17.4\%          & 0.128          \\
CLIP centroid         & 0.6414          & 15.2\%          & 0.171          \\
CLIP nearest member   & 0.6480          & 17.8\%          & 0.170          \\
DINOv2 nearest member   & 0.6468          & \textbf{18.9\%} & \textbf{0.180} \\
DINOv2 centroid         & \textbf{0.6511} & 18.2\%          & 0.179          \\
Oracle                & 0.7932          & 100.0\%         & 1.000          \\
\hline
\end{tabular}
\caption{Attribution performance for $K=12$ using clusters defined in CLIP embedding space.}
\label{tab:rae_clip_defined_clusters}
\end{table}

 \section{Discussion}
This work provides one of the first systematic studies of training-data attribution in flow-matching models. The proposed closed-form method has a strong attribution signal, and does not require retraining or fine-tuning, enabling easy adaptation to dynamically defined clusters. Compared to static-similarity methods based on DINOv2, it performed only slightly worse. The experiments revealed multiple interesting observations. First, static similarity-based attribution methods, especially DINOv2, are very competitive, despite lacking the trajectory-wise knowledge that the analytical closed-form has. Second, attribution highly depends on the latent space representation: RAE, which is based on the DINOv2 autoencoder, achieved considerably better attribution results. VAEs are optimized for reconstruction, indicating distances and similarities are less meaningful in this space. Third, rankings obtained by models trained on different seeds varied, indicating the counterfactual is not only affected by the training samples, but also training stochasticity. This also implies that disagreement is not necessarily a failure of the attribution method.
Attribution remains challenging to define and measure unambiguously, as are similarity criteria. This is also compounded by the stability of flow matching models observed by \citet{briq2026exploring}, suggesting substantial data removal is required in order to alter the model output. It might also partly explain the small separation in performance across various attribution methods.
 
We selected $K=12$ after analyzing the similarity between clusters using multiple $K$ values. When $K$ is large, the clusters are not distinct enough since CelebA-HQ lacks diversity. In this case, attribution might be ambiguous or insensitive as to what cluster is most influential. We experimented with human faces, which are semantically and visually interpretable. Our results generalize to two latent representations and two different architectures, but our experiments are limited to the image domain and large group deletion. Furthermore, the closed-form solution is only available for rectified flow with linear flow paths. Therefore, the questions whether similar behavior persists for other modalities, more fine-grained sample attribution or alternative flow paths remain open, and we aim to study them in follow-up work.


\clearpage
\appendix

\bibliographystyle{unsrtnat}
\bibliography{references}

\end{document}